\documentclass[10pt,twocolumn,letterpaper]{article}

\usepackage{iccv}
\usepackage{times}
\usepackage{epsfig}
\usepackage{graphicx}
\usepackage{amsmath}
\usepackage{amssymb}
\usepackage{multirow}

\usepackage[pagebackref=true,breaklinks=true,letterpaper=true,colorlinks,bookmarks=false]{hyperref}

 \iccvfinalcopy 

\def\iccvPaperID{3632} 

\ificcvfinal\fi
\begin{document}

\definecolor{VangelisColor}{rgb}{0.9,0,0.0} 
\newcommand{\vangelis}[1]{{\color{VangelisColor} Vangelis: #1}}

\definecolor{YangColor}{rgb}{0.0,0.0,0.9} 
\newcommand{\yang}[1]{{\color{YangColor} Yang: #1}}

\definecolor{revcolor}{rgb}{0.5,0.0,0.9}
\newcommand{\rev}[1]{{ #1}}

\newenvironment{my_itemize}{
\begin{description} 
  \setlength{\itemsep}{1pt}
  \setlength{\parskip}{0pt}
  \setlength{\parsep}{0pt}
  }
{\end{description}}

\newcounter{mycounter}
\newenvironment{noindlist}
 {\begin{list}{\arabic{mycounter}.~~}{\usecounter{mycounter} \labelsep=0em \labelwidth=0em \leftmargin=0em \itemindent=0em}}
 {\end{list}}
 
\newcommand{\ba}{\mathbf{a}}
\newcommand{\bb}{\mathbf{b}}
\newcommand{\bc}{\mathbf{c}}
\newcommand{\bd}{\mathbf{d}}
\newcommand{\be}{\mathbf{e}}
\newcommand{\bff}{\mathbf{f}}
\newcommand{\bg}{\mathbf{g}}
\newcommand{\bh}{\mathbf{h}}
\newcommand{\bi}{\mathbf{i}}
\newcommand{\bj}{\mathbf{j}}
\newcommand{\bk}{\mathbf{k}}
\newcommand{\bl}{\mathbf{l}}
\newcommand{\bm}{\mathbf{m}}
\newcommand{\bn}{\mathbf{n}}
\newcommand{\bo}{\mathbf{o}}
\newcommand{\bp}{\mathbf{p}}
\newcommand{\bq}{\mathbf{q}}
\newcommand{\br}{\mathbf{r}}
\newcommand{\bs}{\mathbf{s}}
\newcommand{\bt}{\mathbf{t}}
\newcommand{\bu}{\mathbf{u}}
\newcommand{\bv}{\mathbf{v}}
\newcommand{\bw}{\mathbf{w}}
\newcommand{\bx}{\mathbf{x}}
\newcommand{\by}{\mathbf{y}}
\newcommand{\bz}{\mathbf{z}}
\newcommand{\bA}{\mathbf{A}}
\newcommand{\bB}{\mathbf{B}}
\newcommand{\bC}{\mathbf{C}}
\newcommand{\bD}{\mathbf{D}}
\newcommand{\bE}{\mathbf{E}}
\newcommand{\bF}{\mathbf{F}}
\newcommand{\bG}{\mathbf{G}}
\newcommand{\bH}{\mathbf{H}}
\newcommand{\bI}{\mathbf{I}}
\newcommand{\bJ}{\mathbf{J}}
\newcommand{\bK}{\mathbf{K}}
\newcommand{\bL}{\mathbf{L}}
\newcommand{\bM}{\mathbf{M}}
\newcommand{\bN}{\mathbf{N}}
\newcommand{\bO}{\mathbf{O}}
\newcommand{\bP}{\mathbf{P}}
\newcommand{\bQ}{\mathbf{Q}}
\newcommand{\bR}{\mathbf{R}}
\newcommand{\bS}{\mathbf{S}}
\newcommand{\bT}{\mathbf{T}}
\newcommand{\bU}{\mathbf{U}}
\newcommand{\bV}{\mathbf{V}}
\newcommand{\bW}{\mathbf{W}}
\newcommand{\bX}{\mathbf{X}}
\newcommand{\bY}{\mathbf{Y}}
\newcommand{\bZ}{\mathbf{Z}}
\newcommand{\balpha}{\mbox{\boldmath$\alpha$}}
\newcommand{\bgamma}{\mbox{\boldmath$\gamma$}}
\newcommand{\bGamma}{\mbox{\boldmath$\Gamma$}}
\newcommand{\bmu}{\mbox{\boldmath$\mu$}}
\newcommand{\bphi}{\mbox{\boldmath$\phi$}}
\newcommand{\bPhi}{\mbox{\boldmath$\Phi$}}
\newcommand{\bSigma}{\mbox{\boldmath$\Sigma$}}
\newcommand{\bsigma}{\mbox{\boldmath$\sigma$}}
\newcommand{\btheta}{\mbox{\boldmath$\theta$}}

\newcommand{\mL}{\mathcal{L}}
\newcommand{\mU}{\mathcal{U}}
\newcommand{\mC}{\mathcal{C}}
\newcommand{\mS}{\mathcal{S}}
\newcommand{\mR}{\mathcal{R}}
\newcommand{\mD}{\mathcal{D}}
\newcommand{\mT}{\mathcal{T}}
\newcommand{\mSl}{\mathcal{S}_l}
\newcommand{\mN}{\mathcal{N}}
\newcommand{\mDll}{\mathcal{D}_{l,l'}}

\newcommand{\ra}{\rightarrow}
\newcommand{\la}{\leftarrow}

\def\A{{\cal A}}
\def\B{{\cal B}}
\def\C{{\cal C}}
\def\D{{\cal D}}
\def\E{{\cal E}}
\def\F{{\cal F}}
\def\G{{\cal G}}
\def\H{{\cal H}}
\def\I{{\cal I}}
\def\J{{\cal J}}
\def\K{{\cal K}}
\def\L{{\cal L}}
\def\M{{\cal M}}
\def\N{{\cal N}}
\def\O{{\cal O}}
\def\P{{\cal P}}
\def\Q{{\cal Q}}
\def\R{{\cal R}}
\def\S{{\cal S}}
\def\T{{\cal T}}
\def\U{{\cal U}}
\def\V{{\cal V}}
\def\W{{\cal W}}
\def\X{{\cal X}}
\def\Y{{\cal Y}}
\def\Z{{\cal Z}}
\def\Re{{\mathbb R}}
\def\Cx{{\mathbb C}}
\def\Ze{{\mathbb Z}}
\def\Na{{\mathbb N}}
\def\ud{\mathrm{d}}
\def\eps{\varepsilon}
\def\dist{\textrm{dist}}


\linespread{1}


%

\title{ SceneGraphNet: Neural Message Passing for  3D Indoor Scene  Augmentation \vspace{-2mm}
 }

\author{
Yang Zhou \qquad Zachary While \qquad Evangelos Kalogerakis\\University of Massachusetts, Amherst\\
{\tt\small \{yangzhou,zwhile,kalo\}@cs.umass.edu}
\vspace{-2mm}
}

\maketitle

\begin{abstract}
In this paper we propose a neural message passing approach to augment an input 3D\ indoor scene  with   new objects matching their surroundings. Given an input, potentially incomplete, 3D scene and a query location (Figure \ref{fig:teaser}), our method predicts a probability distribution over object types that fit well in that location.
Our distribution is predicted though passing learned messages in a dense graph whose nodes represent objects in the input scene and edges represent  spatial and structural relationships. By  weighting messages through an  attention mechanism, our method learns to focus on the most relevant surrounding scene context to predict new scene objects. We found that our method significantly outperforms   state-of-the-art approaches  in terms of correctly predicting   objects missing in a scene
based on our experiments in the SUNCG dataset.
We also demonstrate other applications of our method, including context-based 3D\ object recognition and iterative scene generation.

\end{abstract}

\vspace{-3mm}
\section{Introduction}
\begin{figure}[!t]
\begin{center}
\includegraphics[trim={2cm 7.5cm 2cm 7.5cm},clip,width=0.48\textwidth]{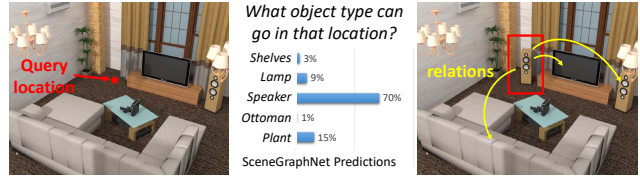}
\end{center}
\vskip -3mm
   \caption{SceneGraphNet  captures relationships between objects in an input 3D\ scene through iterative message passing in a dense graph  to make object type predictions at query locations. }
\vskip -5mm   
\label{fig:teaser}
\end{figure}

With the increasing number of 3D models and scenes becoming available in online repositories, the need for effectively answering object queries in 3D scenes has become greater than ever. A common type of scene queries is to predict plausible object types that match well the surrounding context of an input  3D scene.     For example, as shown in Figure \ref{fig:teaser}, given a query location close  to a TV stand and  a room corner, a likely choice for an object to be added  in that location   can be a speaker, or less likely a plant.

\begin{figure}[t!]
\begin{center}
\includegraphics[trim={2cm 7.0cm 2cm 7.3cm},clip,width=0.48\textwidth]{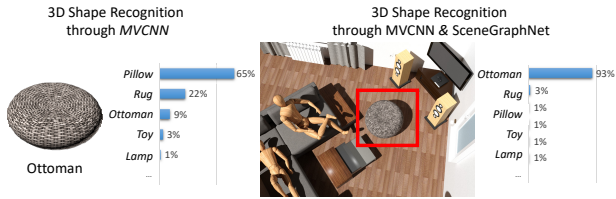}
\end{center}
\vskip -3mm
   \caption{Context-based object recognition. \textbf{Left: }Object recognition using a  multi-view CNN \cite{su15mvcnn} without considering the scene context. \textbf{Right: }Improved recognition by fusing the multi-view CNN and SceneGraphNet predictions based on scene context.}
\vskip -3mm
\label{fig:context_recognition}
\end{figure}

\begin{figure}[t!]
\begin{center}
\includegraphics[trim={5cm 9.1cm 5cm 7.5cm},clip,width=0.5\textwidth]{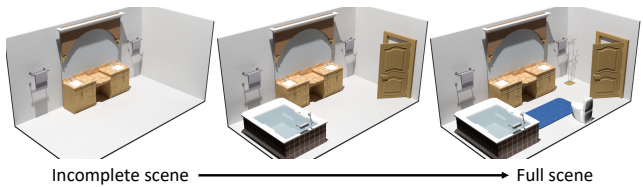}
\end{center}
\vskip -3mm
   \caption{ Iterative scene synthesis. Given an incomplete scene, our method is used to populate it progressively with more objects at their most likely  locations predicted from SceneGraphNet.}
\label{fig:auto-complete}
\vskip -3mm
\end{figure}

We designed a neural message passing method to predict a probability distribution over object types given query locations in  the scene. Our predicted distributions can be used in various vision and graphics tasks. First, our method can enhance 3D object recognition in scenes by taking into account the scene context (Figure \ref{fig:context_recognition}). Second, it can also be used to automatically populate 3D scenes with more objects by evaluating our probability distribution at different  locations in the scene (Figure \ref{fig:auto-complete}). Another related application is to provide object type recommendations to designers while interactively modeling a 3D scenes.

Our method models the scene as a graph, where nodes represent existing objects and edges represent various spatial and structural relationships between them, such as supporting, surrounding, adjacency, and object co-occurrence relationships. The edges are not limited only to neighboring objects, but  can also capture long-range dependencies in the scene e.g., the choice of a sofa in one side of a room can influence the selection of other sofas, chairs, or tables in the opposite side of the room to maintain a plausible object set and arrangement.

Our method is  inspired by graph neural network approaches that learn  message passing in graphs \cite{Battaglia:2016:PYS,Gilmer:2017:MPNNC, Hamilton:2017:GraphSAGE} to infer node representations and interactions between them. Our method  learns
message passing to aggregate the surrounding scene context from different objects. It addresses a number of challenges in this setting. First, scene objects may have multiple types of relationships between them e.g., a nightstand can be adjacent to a bed, while at the same time it is placed symmetrically wrt to another nightstand surrounding the bed from both sides. We found that object relationships are more effectively captured through neural network modules specialized for each type of relationship. In addition, we found that predictions are more plausible, when we model not only local or strictly hierarchical object relationships, but also long-range relationships captured in a dense graph. Since we do not know a priori which relationships are most important for predicting  objects at query locations, we designed an attention mechanism to weigh different messages i.e., we find which edges are more important for making object predictions. Finally, we found that aggregating messages from multiple objects are better handled with  memory-enabled units (GRUs) rather than other simpler schemes, such as summation or max-pooling. 

We tested our method against several alternatives in a large dataset based on SUNCG. We evaluated how accurately different methods predict object types that are intentionally left out from SUNCG scenes. Our method improves prediction of missing object types by a large margin of $16\%$ ($51\% \ra 67\%$) compared to the previous best method adopted for this task.
Our contribution is two-fold:
\begin{itemize}
  \item \vspace{-2mm} a new graph neural network architecture to model short- and long-range relationships between objects in 3D indoor scenes.
  \item \vspace{-2mm} an iterative message passing scheme, reinforced with an attention mechanism, to perform object-related prediction tasks in scenes, including spatial query answering, context-based object recognition, and iterative scene synthesis.
\end{itemize}

\vspace{-2mm}
\section{Related Work}
\begin{figure*}
\begin{center}
\includegraphics[trim={0.1cm 7.25cm 0.1cm 7.1cm},clip,width=0.98\textwidth]{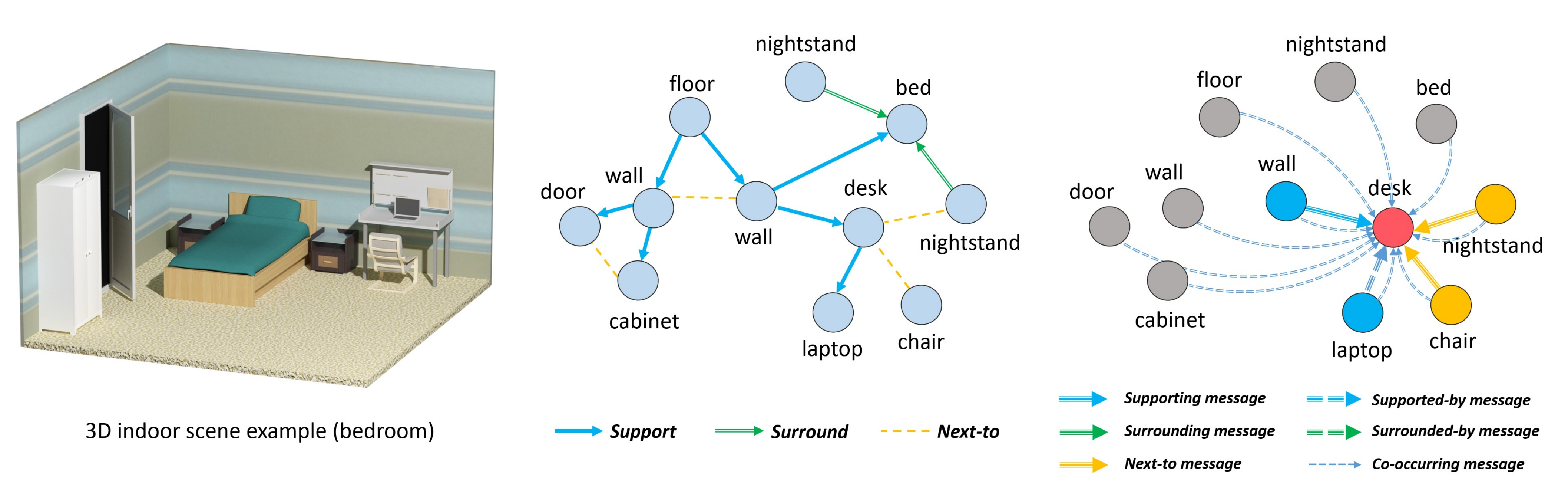}
\end{center}
\vskip -5mm
   \caption{An example of the  graph  structure used for neural message passing in SceneGraphNet for a bedroom scene. \textbf{Left:} an input 3D scene. \textbf{Middle:} graph structure and  object relationships we modeled (some relationships, e.g. dense ``co-occurrence'' and ``next-to'' ones, are skipped for clarity). \textbf{Right:} messages received by the object ``desk'' from  other nodes in the graph for all different types of relationships.}
\label{fig:graph_archi}
\vskip -3mm
\end{figure*}

Our work is  related to learning methods for synthesizing indoor 3D scenes
by  predicting new objects to place in them
either unconditionally or  conditioned on some input, such as spatial queries. 
Our work is also related to  graph neural networks and  neural message passing.

\vspace{-3mm}
\paragraph{Indoor scene synthesis.} Early work in 3D scene modeling employed hand-engineered kernels and graph walks to retrieve objects from a database of 3D models that are compatible with the surrounding context of an input scene 
\cite{Fisher:2010:CSM,Fisher:2011:CSR}.  Alternatively,   Bayesian Networks have been proposed to synthesize arrangements of 3D objects by modeling object co-occurence and placement statistics \cite{Fisher:2012:ESO}. Other  probabilistic graphical models have also been used to model pairwise compatibilities between objects to synthesize scenes given input sketches of scenes \cite{Xu:2013:SSC} or RGBD data  \cite{Chen:2014:ASM,Kermani:2015:LSS}. These earlier methods were mostly limited to small scenes. Their generalization ability was  limited due to the coarse scene relationships,  shape representations, and statistics captured in their shallow or hand-engineered statistical models.  

With  the  availability  of  large  scene  datasets  such  as SUNCG  \cite{song2016ssc}, more sophisticated learning methods have been proposed. Henderson et al. \cite{Henderson:2017:GMO} proposed a Dirichlet process mixture model to model higher-order relationships of objects in terms of co-occurences and relative displacements.
More related to our approach are  deep  networks proposed for scene synthesis and augmentation. Wang et al. \cite{Wang:2018:DCP, Daniel:2018:ImprovedDeepSynth} use image-based CNNs to encode top-down views of  input scenes,  then decode them towards object category and location predictions. In a concurrent work, Zhang et al. \cite{Zhang:2018:DGM} use a Variational Auto-Encoder coupled with a Generative Adversarial Network to generate scenes represented in a matrix where each column represents an object with location and geometry attributes. Most relevant to our work is GRAINS \cite{Li:2019:GGR}, a recursive auto-encoder network that generates scenes represented as tree-structured scene graphs, where internal nodes represent groupings of objects according to various relationships (e.g., surrounding, supporting), and leaves represent object categories and sizes.  GRAINS directly encodes  only local dependencies between objects of the same group in its tree-structured architecture. The tree representing the scene is created through hand-engineered heuristic grouping operations. \rev{Concurrently to our work, Wang et al. \cite{Wang:2019:PPI} proposed a graph neural network for scene synthesis. The edges in the graph represent spatial and semantic relationships of objects, however, they are pruned through heuristics.
Our method instead models scenes as dense graphs capturing both short- and long-range dependencies between objects. Instead of hand-engineering priorities for object relationships, our methods learns to attend the most relevant relationships to augment a scene}.



\vspace{-3mm}
\paragraph{Graph Neural Networks.} A\ significant number of methods has been proposed to model graphs as neural networks \cite{Scarselli2009,Hamilton:2017:survey, Hamilton:2017:GraphSAGE, Scarselli:2009:GRP, Li:2015:Gated, Gilmer:2017:MPNNC, Battaglia:2016:PYS, Schutt:2017:quantum}. Our method is mostly related to approaches that perform message passing along edges to update node representations through neural network operations \cite{Battaglia:2016:PYS,Gilmer:2017:MPNNC, Hamilton:2017:GraphSAGE}. In our graph network, nodes are connected with multiple edges allowing the exchange of information over multiple structural relationships across objects, and  we also use an attention mechanism that weigh the most relevant messages for scene object predictions. We also adapt the graph structure, node representations,   message ordering and aggregation  in the particular setting of 3D\ indoor scene modeling.








\vspace{-0mm}
\section{Method}

The input to our method is a set of 3D  models of objects arranged in a scene $\bs$. We assume that the current objects in the scene are labeled based on their types (e.g., sofa, table, and so on).  Given a query location $\bp$ in the scene (Figure \ref{fig:teaser}), the output of our method is a probability distribution over different object types, or categories, $P( C | \bp,\bs)$ expressing how likely is for objects from each of these categories to fit well in this location and match the scene context. The probability distribution can be used in various ways, such as simply selecting an object from a category with the highest probability to be placed in the query location for scene augmentation tasks, or presenting a list of object type recommendations to a designer ordered by their probability for interactive scene modeling tasks. Alternatively, for object recognition tasks our distribution can be combined with a posterior that attempts to predict the object category based only on individual shape data.

To determine the target probability distribution, our method first creates a graph whose nodes represent objects in the scene and edges represent different types of relationships between objects (Figure \ref{fig:graph_archi}). Information flows in this graph by iteratively passing learned messages between nodes connected through edges.
We note that the graph we use  for message passing should not be confused
with scene graphs (often in the form of trees)\ used in graphics file formats for representing objects in scenes based on   hierarchical transformations. Our scene graph representation has a much richer structure and is not limited to a tree.  

In the following sections, we explain message passing  (Section \ref{sec:message_passing}),  different strategies for  designing the graph structure (Section \ref{sec:graph_structure}), the target distribution prediction (Section \ref{sec:prediction}), and   applications (Section \ref{sec:application}). 

\subsection{Message passing}
\label{sec:message_passing}

\begin{figure*}
\begin{center}
\includegraphics[trim={2.0cm 3.08cm 1.9cm 3.3cm},clip,width=0.98\textwidth]{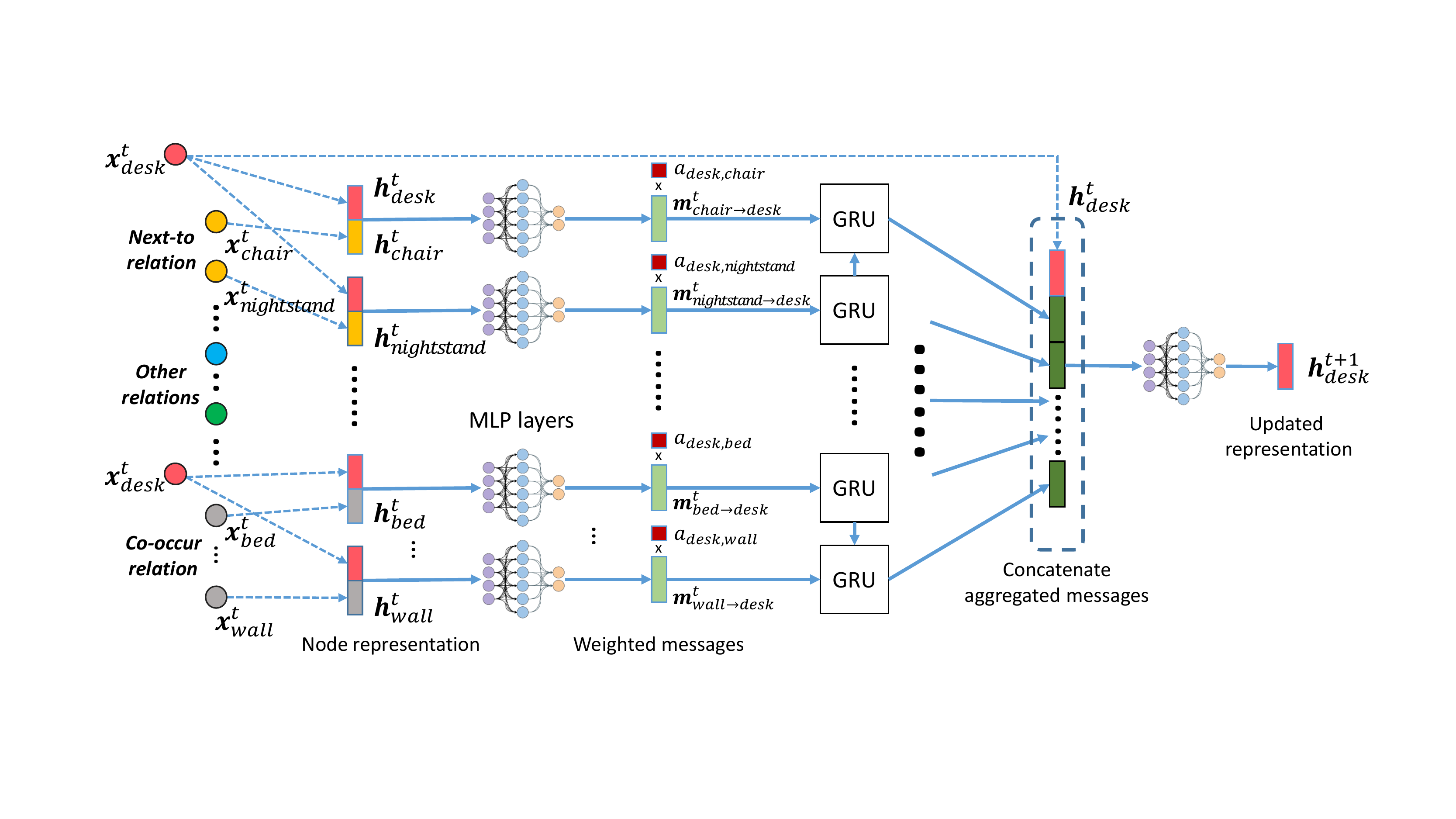}
\end{center}
\vskip -7mm
   \caption{Overview of our message passing and underlying neural network architecture. We take the example in Figure \ref{fig:graph_archi} to illustrate a single message passing iteration.}
\vskip -3mm
\label{fig:network_flow}
\end{figure*}

Figure \ref{fig:network_flow} illustrates our message passing and underlying neural architecture.
Each node $i$ in our scene graph represents a 3D object. The node  internally carries a  vectorial representation  $\bh_i$ encoding information about both the shape and also its scene context based on the messages  it receives from other nodes.  The messages are used to update the node representations so that these reflect the  scene context captured in these messages. New messages are emitted from nodes, which result in more information exchange about scene context.
In this manner, the message passing procedure  runs iteratively. In the following paragraphs, we describe the steps of this message passing procedure.  

\textbf{Initialization.} Each node representation is initialized based on the shape representation $\bx_i$ at the node:
\begin{equation}
\bh_i^{(0)} = f_{init}( \bx_i; \bw_{init})
\end{equation}
where $f_{init}$ is a two-layer MLP  with learnable parameters $\bw_{init}$  outputting a $100$-dimensional node representation (details are provided in the supplementary material). In our implementation, the shape representation is formed by the concatenation of three vectors $\bx_i = [ \bc_i, \bp_i, \bd_i]$ where $\bc_i$ is a one-hot vector representing its category, $\bp_i \in \mR^3$ is its centroid 3D position in the scene, $\bd_i \in \mR^3$ is its scale (its oriented bounding box lengths).

\textbf{Messages.} A message from a node $k$ to node $i$ encodes an exchange of information, or interaction, between their corresponding shapes. The message carries information based on the corresponding node representations $\bh_i^{(t)}$ and $\bh_k^{(t)}$ and also depends on the type of relationship $r$ between the two nodes (e.g., a different message is computed for a ``surrounding'' relationship or for a ``supporting'' relationship). We discuss  different types of relationships in Section \ref{sec:graph_structure}. At step $t$ the message $\bm_{k \ra i}^{(r,t)}$  from node $k$ to another node $i$ is computed as:
\begin{equation}
\bm_{k \ra i}^{(r,t)} = f_{msg}^{(r)}( \bh_k^{(t)}, \bh_i^{(t)}; \bw_{msg}^{(r)}) 
\end{equation}
where $f_{msg}^{(r)}$ is a two-layer MLP with learnable parameters $\bw_{msg}^{(r)}$ (weights are different per relationship $r$) outputting a 100-dimensional message representation.
 
\textbf{Weights on messages.} Some messages   might be more important, or more relevant for prediction, than others. Thus, we implemented a form of   attention mechanism for messages. During message passing, each message $\bm_{k \ra i}^{(r,t)}$ is scaled (multiplied) with a scalar weight $a_{k,i}$ computed as follows:
\begin{equation}
a_{k,i} = f_{att}( \bx_k, \bx_i ; \bw_{att} ) 
\end{equation}
where $f_{att}$ is a two-layer MLP followed  by a sigmoid layer with learnable parameters $\bw_{att}$. Note that the attention weights are computed from the raw shape representations.  We practically found that this strategy had more stable behavior (better  convergence) compared to updating them using both the shape and latent representations. Furthermore, weights that are almost  $0$ imply that the interaction between two nodes is negligible. This can be used in turn  to discard edges  and accelerate message passing at test time without sacrificing performance.  Note also that the shape representations include positions of objects, thus weights are expected to correlate with  object distances. 

\textbf{Message aggregation.} Messages are aggregated through GRU modules learned for each type of relationship. Specifically, for each node $i$, we pass a sequence of messages $\{ \bm_{k \ra i}^{(r,t)} \}_{k \in \mN(i)}$ coming from all the different nodes $k \in \mN(i)$ connected to it   through a series of GRU units (where $\mN(i)$\ represents the set of  nodes emitting to node $i$). Each GRU unit receives as input a message from an emitting node $k$ and the previous GRU\ unit in the sequence, and produces an aggregated message $\bg_{i,j}^{(r,t)}$ ($j=1,...,|N(i)|$  is an index for each GRU unit) as follows:
\begin{equation}
\bg_{i,j}^{(r,t)} = f_{GRU}^{(r)}( \bg_{i,j-1}^{(r,t)}, a_{k,i} \cdot \bm_{k \ra i}^{(r,t)};  \bw_{GRU}^{(r)})
\end{equation}
where $\bw_{GRU}^{(r)}$ are learnable GRU\ parameters per type of relationship $r$. The last GRU unit in the sequence produces the final aggregated message, denoted simply as $\bg_{i}^{(r,t)}$ (dropping the last index $j=|\mN(i)|$ in the GRU sequence for clarity). We note that the first GRU\ unit in the sequence receives an all-zero message.
 The order of messages also matters, as generally observed   in recurrent networks. In our implementation, the messages from emitting nodes to the node $i$ are ordered according to the Euclidean distance
of their corresponding object centroid  to the centroid of the object represented at node $i$ (from furthest to closest). 

\textbf{Node representation update.} Finally, for each node $i$, its latent representation is updated to reflect the transmitted context captured in all messages sent to it across all different types of relationships. Specifically, the aggregated messages $\bg_{i}^{(r,t)}$ from all relationship types are concatenated,  and its latent representation is updated as follows:
\begin{equation}
\bh_i^{(t+1)} = f_{upd}( \bh_i^{(t)}, Concat(\{ \bg_i^{r,t} \}_{r \in \mR}); \bw_{upd})
\end{equation}
where $f_{upd}$ is a two-layer MLP with learnable weights $\bw_{upd}$, and $\mR$ is the set of relationships that we discuss in the following section.

\subsection{Graph structure and Object Relationships}
\label{sec:graph_structure}
\label{sec:structure_graph}

A crucial component in our method is the underlying graph structure used for message passing. One obvious strategy is to connect each node
(object)
 in a scene with  neighboring nodes (objects) based e.g., on Euclidean distance between them. However, we found that connecting the graph via simple  neighboring relationships resulted in poor object prediction performance  for a number of reasons. First, it is hard to define a global distance threshold or single number of nearest neighbors that work  well for all different scenes. Second,  and 
most importantly, such ``neighboring'' relationships are coarse and often ambiguous. For example, a pillow and a  nightstand are both ``nearby''  a bed,
yet
their structural and spatial relationships wrt the bed can be quite different i.e., the pillow is ``on top of'' the bed, while a nightstand is ``adjacent'' or ``next to'' the bed. We found that representing more fine-grained  relationships in the graph  significantly increases the prediction  performance of our method.  We also found that forming dense graphs capturing long-range interactions between scene objects was better than using sparse graphs or constraining them to be tree-structured, as discussed in our results section. Below we discuss the different types of relationships we used to form connections between nodes for message passing.

\vspace{-4mm}
\paragraph{``Supporting'' relationship.} A node $i$ is connected to a node $k$ via a directed edge of
this relationship if the object represented by node $i$ supports, or is ``on top of'', the object at node $k$. The relationship can be detected by examining the bounding boxes of the two objects (details are provided in the supplementary material). 

\vspace{-4mm}
\paragraph{``Supported-by'' relationship.} This is the opposite relationship to the previous one i.e.,  a node $i$ is connected to a node $k$ via an edge of this relationship if the object at node $i$ is  supported by node $k$. Note that a ``supporting'' relationship $i \ra k $ implies  a ``supported-by' relationship $k \ra i$.  Yet, we use different weights for these relationships i.e., the node $i$ sends a message to $k$ learned through a   MLP whose weights are different from the MLP\ used to infer the message from $k$ to $i$. Using such assymetric messages improved performance compared to using symmetric ones, since in this manner  we capture  the directionality of this relationship. Having exclusively ``supporting'' relationships (and not vice versa) also resulted in lower performance.

\vspace{-4mm}
\paragraph{``Surrounding'' relationship. } If there is a set of objects surrounding another object  i.e., the set has objects of same size whose bounding boxes are placed under a reflective or rotational symmetry around a central object, then all these objects are connected to the central object via directed edges of the ``surrounding''  relationship type (details are provided in the supplementary material). 

\vspace{-4mm}
\paragraph{``Surrounded-by'' relationship. } This is the opposite relationship to the previous one. A central object is connected to the objects surrounding it via directed edges of `surrounded-by''  relationship type. 

\vspace{-4mm}   
\paragraph{`Next-to'' relationship.} Two nodes $i$ and $k$ are connected  via an undirected edge of ``next-to'' relationship type if the node $i$ is adjacent to node $k$ and lie on the same underlying supporting surface. Note that in contrast to the previous relationships,  this is a symmetric relationship, thus messages are inferred by the same MLP\ in  both directions. 

\vspace{-4mm}
\paragraph{``Co-occuring'' relationship.} Two nodes $i$ and $k$ are connected  via an undirected edge of ``co-occuring'' relationship type if their objects co-exist in the same  scene. This type of relationship  obviously results in a fully connected graph. As discussed earlier in the introduction, the selection of objects to place in a location may be strongly influenced by other objects further away from that location. 
We found that capturing such long-range, ``object co-occurence'' interactions, in combination with our attention mechanism that learns to weigh them, improved the prediction performance of our method, even if this happens at the expense of more computation time. We note that edges can be dynamically discarded during message passing by examining the attention weights in the graph, thus accelerating the execution at test time. For scenes containing 50-100 objects, message passing  in our graph takes  a few seconds at test time. 

Figure \ref{fig:graph_archi} illustrates our scene graph structure with all the different types of relationships for a toy scene example. We note that in all our training and test scenes, there is a ``floor'' node and ``wall'' nodes, since these objects are also constituent parts of indoor scenes and are  useful for making predictions (i.e., when we want to predict an object hanging on the wall).

\subsection{Prediction}
\label{sec:prediction}

Given a query location in a form of a point $\bp$ in the scene, we form a special ``empty'' node $m$  in our graph
representing the ``missing object'' to predict. The node is
 initialized to an all-zero shape category and size representation vector, and 3D\ position set to the $\bp$. We connect it to other nodes in the graph based on the relationships discussed in the previous section (including our directed relationships which can be inferred by examining the relative position of the query point wrt other objects in the scene). This special node along with its edges form our final scene graph $\bs$ based on the input scene and query. 

 We then execute message passing in our graph. The node representations are updated at each time step synchronously (i.e.,  all messages  are computed before they are sent to other nodes), including the representation $\bh_m^{(t)}$
of our special node. Message passing is executed up to $t=T$ time steps (practically $T=3$ iterations in our implementation). Then the representation in our special node is decoded through a two-layer MLP and a softmax to predict our target probability distribution: 
\begin{equation}
P(\bc | \bp,\bs) = f_{pred}( \bh_m^{(T)}; \bw_{pred} )
\end{equation}
where $f_{pred}$ represents the MLP and $\bw_{pred}$ its learnable parameters. For interactive modeling tasks, we also found useful to predict the size $\bd_m$ of the object to place in the scene. This is done through one more MLP regressing to object size from the learned node representation.

\begin{figure}
\begin{center}
\includegraphics[trim={6.2cm 3.2cm 4.cm 2.0cm},clip,width=0.5\textwidth]{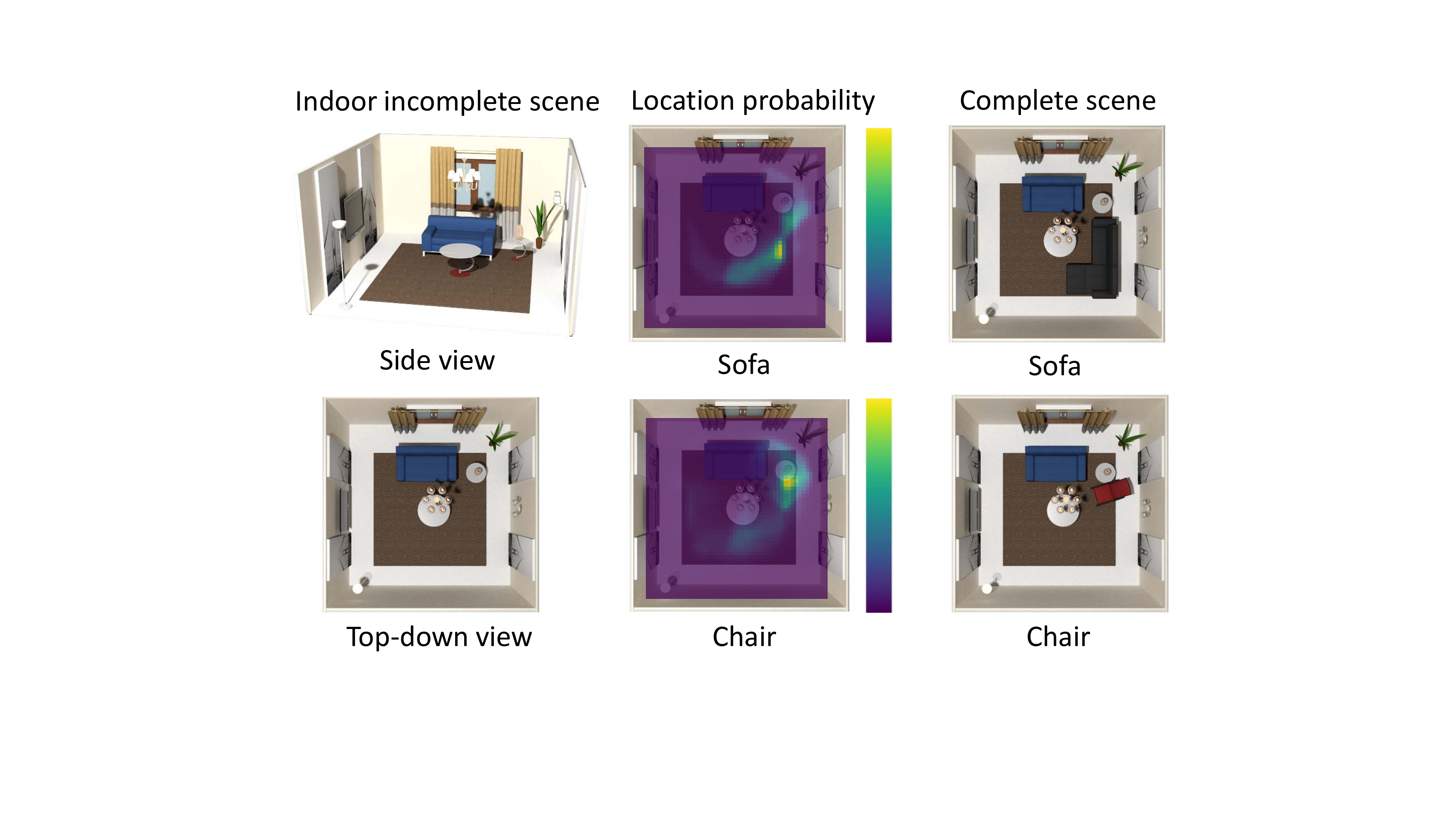}
\end{center}
\vskip -3mm
   \caption{Prediction of most likely object categories to add in the scene  and associated placement distributions.
   \textbf{Left: }An input scene. \textbf{Middle: }the top-two most plausible categories to add in the scene (sofa and chair) along with the evaluated placement probability at each scene location. \textbf{Right:} Resulting scene with a  sofa and chair placed  at the most likely location.}
\vskip -3mm
\label{fig:pos_grid}
\end{figure}

\vspace{-2mm}
\paragraph{Training.} All the MLPs in our network, including the GRU modules, are trained jointly. To train our graph network, given a training scene, we remove a random object from it (excluding ``walls'' and ``floor''). Our training aims to predict correctly the category of the removed object. This means that we place a query in its location, and execute our message passing interactive procedure. Based on the predicted distribution for the corresponding empty node, we form a  categorical cross entropy loss for the missing object's category.
 To train the MLP\ regressing to the object size, we additionally use the $L_2$ loss between  the ground-truth size and the predicted one.


\vspace{-2mm}
\paragraph{Implementation details.} Training is done through the Adam optimizer \cite{Adam} with learning rate $0.001$, beta coefficients are $(0.9, 0.999)$ and weight decay is set to $10^{-5}$. The batch size is set to $350$ scenes. The model converges around 8K iterations for our largest room dataset. We pick the best model and hyper-parameters based on  hold-out validation performance. \rev{Our  implementation is in Pytorch and is provided at:{ \small \url{https://github.com/yzhou359/3DIndoor-SceneGraphNet}} }.

\subsection{Applications}
\label{sec:application}
\vspace{-1mm}
\paragraph{Object recognition in scenes.} Objects in 3D scenes may not be always tagged based on their category. One approach to automatically recognize 3D objects is to process them individually through standard 3D shape processing architectures operating on a volumetric (e.g. \cite{Wu15}), multi-view (e.g., \cite{su15mvcnn}), or point-based representations (e.g., \cite{qi2017pointnet}). However, such approaches are not perfect and are susceptible to mistakes, especially if there are objects whose shapes are similar across different  categories. To improve object recognition in scenes, we can also use the contextual predictions from SceneGraphNet. Specifically, given a posterior distribution $P(C | \bo)$ for an object extracted by one of the above 3D\ deep architectures given its raw shape representation $\bo$   (voxels, multi-view images, or points), and our posterior distribution extracted from the node at the location of the object $P(C | \bp,\bs)$, we can simply take the product of these two distributions and re-normalize. In our experiments, we used a popular multi-view architecture \cite{su15mvcnn}, and   found that the resulting distribution yields better predictions compared to using multi-view object predictions alone. Figure \ref{fig:context_recognition} demonstrates a characteristic example. 

\vspace{-2mm}
\paragraph{Incremental scene synthesis.} Our posterior distribution is conditioned at a query location to evaluate the fitness of an object category at that location. We can use our distribution to incrementally synthesize a scene by evaluating $P(C | \bp,\bs)$  over a grid of query locations $\bp$ in the scene (e.g., a 2D regular grid of locations on the floor, or the surface of a randomly picked object in a scene, such as desk), picking the location and object category that maximizes the distribution, then retrieving
a 3D\ model from a database that matches the predicted size at the location (we note that we assume a uniform prior distribution over locations here).
Figure \ref{fig:pos_grid} shows examples of the  most likely object prediction and the evaluated probability distribution for placement across  the scene.
Figure \ref{fig:auto-complete}
shows an example of  iterative scene synthesis. Although some user supervision is eventually required to tune placement, \rev{specify object orientation, and stop the iterative procedure}, we believe that our method may still provide helpful guidance for 3D scene design.

\vspace{-0mm}
\section{Results and Evaluation}
We evaluated our method both qualitatively and quantitavely. Below we discuss our dataset, evaluation metrics, comparisons with alternatives, and our ablation study. 

\vspace{-3mm}
\paragraph{Dataset.}\label{sec:dataset} Following 
\cite{Wang:2018:DCP}, we experimented with four room types (6K\ bedrooms, 4K\ living rooms, 3K\ bathrooms, and 2K\ offices) from the SUNCG dataset. The number of object categories varies from 31 to 51. Dataset statistics are presented in the supplementary material. To ensure fair comparisons with other methods \cite{Li:2019:GGR, Wang:2018:DCP} that assume rectangular rooms with four walls as input, we excluded  SUNCG\ rooms that do not have this layout (we note that our method does not have this limitation).
\rev{All methods were trained for each of the four room types separately. We used a random split of $80\%$ of scenes for training, $10\%$ for hold-out validation, $10\%$ for testing (the same splits and hyper-parameter tuning procedure were used to train all methods)}.

\vspace{-3mm}
\paragraph{Evaluation metrics.} For the task of scene augmentation conditioned on a 3D query point location, we used the following procedure for evaluation. Given a test SUNCG scene, we randomly remove one of the objects (excluding floor and walls). Then given a query location set to centroid of this object, we compute the object category prediction distributions from all competing methods discussed below. First, we measure the classification accuracy i.e., whether the most likely object category prediction produced by a method agrees with the ground-truth category of the removed object. We also evaluate top-K\ classification accuracy, which measures whether the ground-truth category is included in the K most probable predictions produced by a method. The reason for using the top-K accuracy is that some objects from one category (e.g., a laptop on a desk) can be replaced with objects from another category (e.g., a book) in a scene
without sacrificing the scene plausibility.
Thus, if a method predicts book as the most likely category, then laptop, the top-K accuracy (K\textgreater1) would be unaffected. 
\rev{
We also evaluate the accuracy of  predictions for object size. Objects are annotated with physical size in SUNCG, thus we report the error in terms of centimeters (cm) averaged in all three dimensions across all test queries.
}

\vspace{-4mm}
\paragraph{Comparisons.} We compare with two state-of-the-art methods for scene synthesis: GRAINS \cite{Li:2019:GGR} and Wang et al.'s view-based convolutional prior \cite{Wang:2018:DCP}. Wang et al. \cite{Wang:2018:DCP} trains a CNN\ module that takes as input a top-down view of a scene and outputs object category predictions conditioned on a 2D location in that view. To compare with Wang et al. \cite{Wang:2018:DCP}, we project our input 3D\ query location onto the corresponding 2D\ location in the top-down view of the scene. We use their publicly available code
to train their module. To compare with GRAINS  \cite{Li:2019:GGR}, we first encode the input scene as a tree based on  
its heuristics (again, we use  their publicly available code). For a fair comparison (i.e., provide same input information to GRAINS as in our method), we also include an ``empty'' node representing the query, connecting it to rest of the tree based on the  same procedure as ours. The node has the same  all-zero  category and size representation as in our method, and its 3D position is set according to the query position expressed relatively to its sibling in the tree  (GRAINS encodes relative positions of objects relative to sibling nodes). The tree is processed through their recursive network, then decoded it to the same tree with the goal to predict the category of the ``empty'' node. To train GRAINS, we used the same loss as our method. The same dataset and splits were used for all methods. We note that GRAINs has $5M$ learnable parameters, Wang et al. has $42M$, while SceneGraphNet has much fewer ($1.5M$).

\begin{figure*}
\begin{center}
\includegraphics[trim={0.2cm 0.9cm 0cm 1.25cm},clip,width=0.98\textwidth]{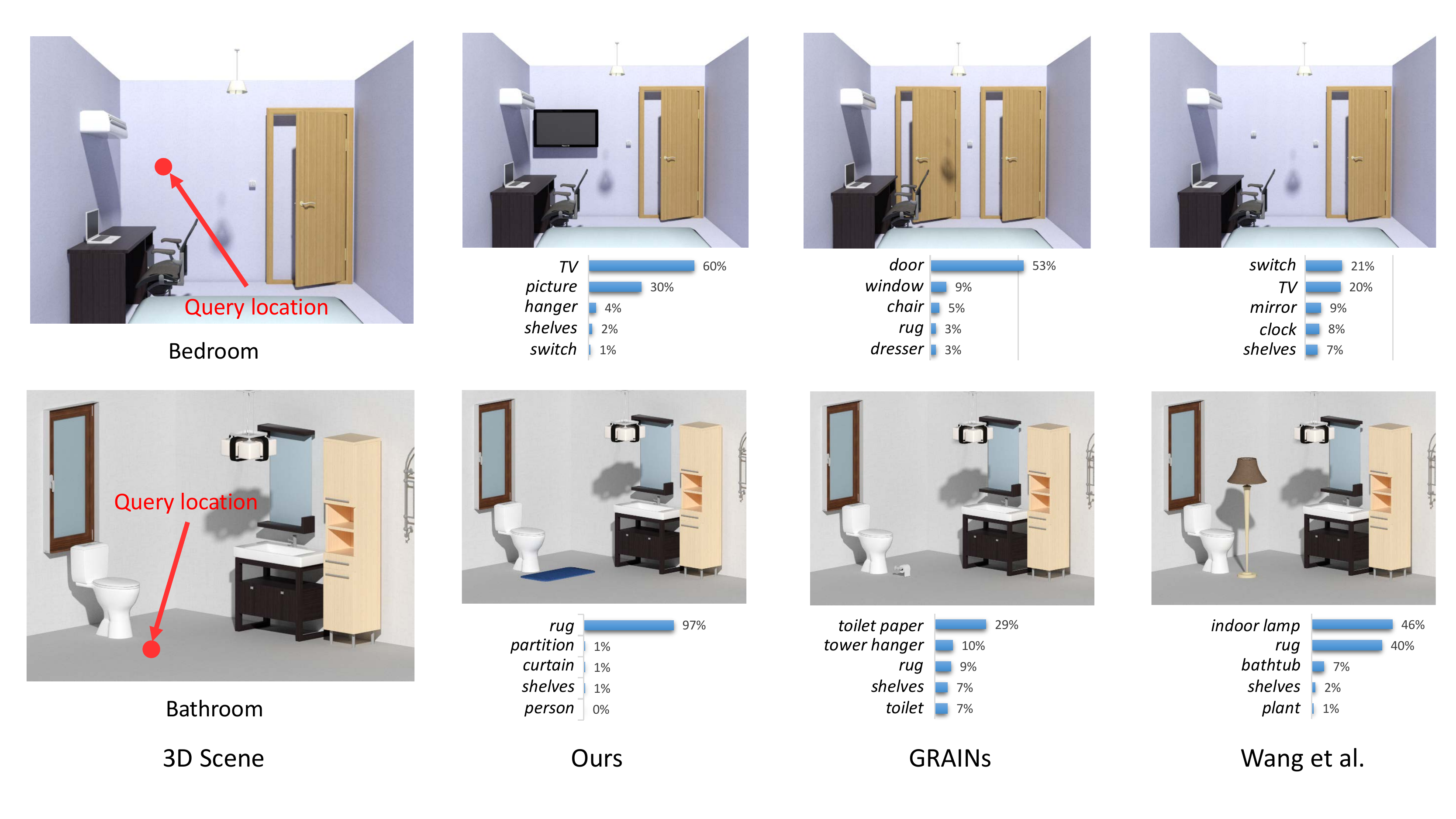}
\end{center}
\vskip -4mm
   \caption{Comparison of  object category predictions for two 3D scenes and\ query positions  (red points)\ across different methods. Given the input  scenes and queries (left), we show the predicted category distributions and  rendered scenes with  objects added from the most likely predicted category per each method.}
\vskip -5mm
\label{fig:top1_render}
\end{figure*}
\begin{figure}
\begin{center}
\includegraphics[trim={4.5cm 7.1cm 5.5cm 6.3cm},clip,width=0.48\textwidth]{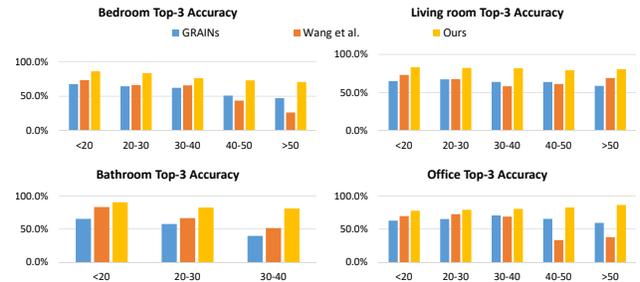}
\end{center}
\vskip -6mm
   \caption{Average top-3 classification accuracy  for rooms  binned according to  number of objects they contain.}
\vskip -5mm
\label{fig:acc_vs_num_object}
\end{figure}

\vspace{-4mm}
\paragraph{Results.} Table \ref{table:scene_completion_avg} shows the top-$K$ accuracy averaged over  the whole dataset across all room types for all  different methods\ $(K=1,3,5$). Our method outperforms the two other competing methods with a significant margin ranging from $10.2\%$ for top-5 accuracy
 to $16.4\%$ for top-1 accuracy.  In the supplementary material, we include detailed evaluation for each room type. Figure \ref{fig:acc_vs_num_object} shows the top-3 accuracy per each room type binned according  to  the total number of objects in the input scene. We observe that the performance gap between our method and others tends to increase  for  larger scenes with more objects.
Figure \ref{fig:top1_render} shows the most likely predicted categories  for SceneGraphNet,  GRAINS \cite{Li:2019:GGR}, and Wang et al. \cite{Wang:2018:DCP} for  two  input scenes and   queries. 

\begin{table}
\begin{center}
\begin{tabular}{|c|c|c|c|}
\hline
\multirow{2}{*}{Method} & \multicolumn{3}{|c|}{Average} \\\cline{2-4}
  & Top1 & Top3 & Top5 \\
\hline\hline
GRAINs\cite{Li:2019:GGR} & 44.2 & 63.9 & 73.6 \\
Wang et al. \cite{Wang:2018:DCP} & 50.9 & 70.7 & 79.3 \\
\cline{1-4}
SGNet-tree & 61.1 & 79.7 & 87.0 \\
SGNet-sparse & 60.0 & 78.6 & 86.5  \\
SGNet-co-occur & 56.5 & 75.4 & 83.3  \\
SGNet-sum & 57.7 & 77.6 & 85.1  \\
SGNet-max & 63.1 & 81.5 & 87.3  \\
SGNet-vanilla-rnn & 64.8 & 82.3 & 88.5  \\
SGNet-no-attention & 60.3 & 79.1 & 85.8  \\
SGNet-dist-weights & 63.8 & 81.6 & 87.9  \\
\cline{1-4}
SceneGraphNet (full model) & \textbf{67.3} & \textbf{83.8} & \textbf{89.5}  \\

\hline
\end{tabular}
\end{center}
\vskip -2mm
\caption{ Top-$K$ accuracy for our scene augmentation task for different methods and variants of SceneGraphNet.}
\vskip -6mm
\label{table:scene_completion_avg}
\end{table}

\rev{
In terms of object size, we compare our method with GRAINS, since GRAINS is also able to predict the oriented bounding box size of the object to add in the scene. We found that GRAINS has $38$cm average error in size prediction, while our method results in much lower error: $26$cm.
}

\vspace{-4mm}
\paragraph{Ablation Study.} We also evaluated our method against several other degraded variants  of it based on the same dataset and classification evaluation metrics (Table \ref{table:scene_completion_avg}). We tested the following variants. 
\textbf{SGNet-tree:} we build the scene graph using only supporting, supported-by, surrounding, and surrounded-by relationships such that the resulting graph has guaranteed tree structure (we break cycles, if any, by prioritizing supporting relationships). \textbf{SGNet-sparse} builds the graph with all the relationships except for the dense ``co-occurring'' ones, resulting in a sparse graph structure.
\textbf{SGNet-co-occur:} we build a fully connected graph with only the ``co-occurring" relationships between edges.  \textbf{SGNet-sum} uses the full graph structure, yet  aggregates messages using a summation over their representations instead of a GRU\ module.
\textbf{SGNet-max} instead aggregates messages using max-pooling over their representations. \textbf{SGNet-vanilla-rnn}  instead aggregates messages using a vanilla RNN. \textbf{SGNet-no-attention} uses the full graph structure, yet does not use the weighting mechanism on messages (all messages have same weight). \textbf{SGNet-dist-weights} instead computes weights on messages by setting them  according to Euclidean distance between nodes (weights are set as $\alpha_{k,i} = c \cdot \exp^{-\frac{||\bd_{k,i}||}{b}}$, where $\bd_{k,i}$ is the Euclidean distance between object centroids $k$ and $i$, $c$ and $b$ are parameters set through hold-out validation). We found that our method outperforms all  these degraded variants.

\vspace{-4mm}
\rev{
\paragraph{Performance wrt iteration number.} After completing the first iteration of message passing, the top-5 accuracy is $72.7\%$. In the second iteration, it increases to $88.2\%$, and in the third one, the performance arrives at $89.5\%$. The performance oscillates around $89\%$ after the third iteration. }

\vspace{-4mm}
\paragraph{Object recognition in 3D scenes.} For each object in our scene dataset, we use a MVCNN \cite{su15mvcnn} to predict its object category $P(C | \bo)$ given its multi-view representation $\bo$. Then we multiply this predicted distribution with the context-aware predicted posterior $P(C | \bp,\bs)$ from SceneGraphNet.
Table \ref{table:prior} shows the average classification accuracy for object recognition with and without combining the SceneGraphNet's posterior with the MVCNN one. Our method significantly improves the accuracy in this context-based object recognition task by  $13\% $. 

\vspace{-4mm}
\paragraph{Timings.} It takes $30$ hours to train our network in our largest  room dataset (bedrooms) with $5K$ training scenes measured on a GeForce 1080Ti GPU. At test time, given an average-sized scene with $40$ objects, our method takes around $0.58$ sec to infer the distribution given the query.

\begin{table}
\begin{center}
\begin{tabular}{|c|c|c|c|c|c|}
\hline
Method & Bed & Living & Bath & Office & Avg \\
\hline\hline
MVCNN & 69.6 & 55.8 & 43.4 & 67.8 & 59.2 \\
MVCNN+ Ours & 79.9 & 74.7 & 56.4 & 73.0 & 72.2 \\
\hline
\end{tabular}
\end{center}
\vskip -2mm
\caption{Object recognition accuracy using MVCNN alone  \cite{su15mvcnn} and using the MVCNN together with our ShapeNetGraph posterior evaluated for the objects in our test scenes.}
\vskip -3mm
\label{table:prior} 
\end{table}

\vspace{-0mm}
\section{Discussion}
We presented a neural message passing method that operates on a dense scene graph representation of a 3D scene to perform scene augmentation and context-based object recognition. There are several avenues for future work. Our method is currently limited to predicting object categories at a given location. It could be extended to instead generate objects and whole scenes from scratch.  \rev{Our message passing update scheme is not guaranteed to converge, nor we forced it to be a contraction map.  Oscillations can happen as in other message passing algorithms e.g., loopy belief propagation. Training and testing with the same number of iterations helps preventing unstable behavior. 
It would be fruitful to investigate modifications that would result in any theoretical convergence guarantees.} Enriching the set of  relationships in our scene graph representation could also help improving the performance. \rev{Another limitation is that our method currently uses the coarse-grained labels provided in SUNCG, as other methods did. Fine-grained and hierarchical classification are also  interesting future directions. }

\vspace{-6mm}
\paragraph{Acknowledgements.} This research is funded by NSF (CHS-161733). Our experiments were performed in the \mbox{UMass} GPU cluster obtained under the Collaborative Fund managed by the Massachusetts Technology Collaborative.

{\small
\bibliographystyle{ieee}
\bibliography{egbib}
}

\clearpage
\section*{Supplementary Materials}

\paragraph{Neural network details.}
The initialization MLP $f_{init}$ takes as input a $(C+6)$-dimensional raw object representation vector, where $C$ is the number of object categories, and the rest $6$ dimensions represent the object's 3D position $\bp \in \mathcal{R}^3$ and scale $\bd \in \mathcal{R}^3$. It processes the input with a hidden  layer of $300$  units, then ReLUs, and another hidden layer of  $300$  units followed by  ReLUs.  It  outputs a $100$-dimensional node representation.

The message MLP $f_{msg}^{r}$ takes as input two concatenated $100$-dimensional node representations.
It processes the input with a hidden  layer of $300$  units, then ReLUs, and another hidden layer of  $300$  units followed by  ReLUs.  It  outputs a $100$-dimensional  message representation. 

The attention network $f_{att}$ takes as input two concatenated raw object representations (input $2C+12$ dimensions).\ It processes the input with a hidden  layer of $300$  units, then ReLUs, and another hidden layer of  $300$  units followed by  a sigmoid activation.  The output is a single scalar weight between $0$ and $1$.

The aggregation network $f_{GRU}$ processes $100$-dimensional message representations at each step. Its internal memory and output is $100$-dimensional. 

The update network $f_{upd}$ operates on the node representation concatenated with the aggregated messages from six relationships
($700-$dimensional input). It processes the input with a hidden  layer of $300$  units, then ReLUs, and another hidden layer of  $300$  units followed by  ReLUs.  It  outputs a $100$-dimensional  updated node representation.

The prediction MLP processes the input $100$-dimensional node representation through a hidden  layer of $300$  units, then ReLUs, and another hidden layer of $300$ units followed by  a softmax activation to output object category probabilities. We also use a MLP\ with the same structure\ to regress to  object size. The output of the last hidden layer  for this MLP is linearly transformed to object size.

\paragraph{Details on relationship extraction.} The ``supporting'' relationship between two objects $(i,j)$ is calculated by checking if the bottom of the object $i$'s bounding box is higher and also within a short range of distance (set as $0.05$ meters to prevent  placement errors) compared to the top surface of the object $j$'s bounding box. The ``surrounding'' relationship is calculated by finding whether objects $(k_1, k_2,...)$ form a symmetry wrt the central object (using a threshold difference of $0.2$ meters). In addition, the surrounding objects should have similar sizes (they qualify as similar if their bounding boxes differ less than a factor $1.2$x when compared to each other).

\begin{figure}
\begin{center}
\includegraphics[trim={9.5cm 8.3cm 9cm 7.6cm},clip,width=0.40\textwidth]{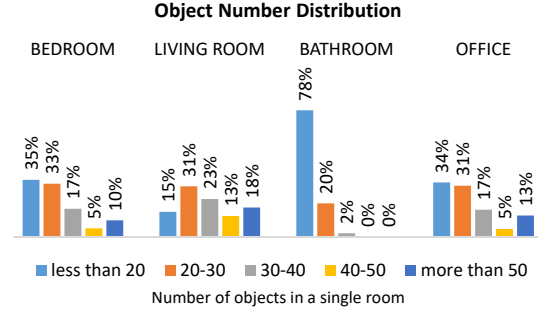}
\end{center}
   \caption{Distribution of \#objects for each room type.}
\label{fig:num_obj_distribution}
\end{figure}

\paragraph{Dataset details.}
Following Wang et al. [20], the experiments are performed on the SUNCG dataset with four room types: bedroom, living room, bathroom and office. We have $51$ object categories in bedrooms, $31$ in bathrooms, $51$ in living rooms, and $42$ in offices. We also count the number of objects in each room per room type.  Figure \ref{fig:num_obj_distribution} shows a histogram over number of objects inside a room per each room type. We note that we will publish the splits and implementation upon acceptance.

\begin{figure}
\begin{center}
\includegraphics[trim={8cm 7.5cm 8cm 6.5cm},clip,width=0.48\textwidth]{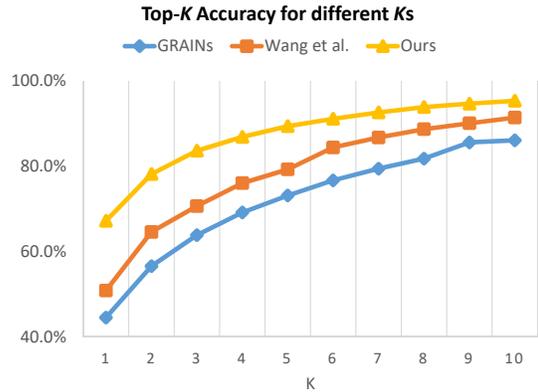}
\end{center}
   \caption{Top-$K$ accuracy of object category prediction for different $K$s.}
\label{fig:topk_performance}
\end{figure}

\begin{table*}[t!]
\small
\begin{center}
\begin{tabular}{|c|c|c|c|c|c|c|c|c|c|c|c|c|}
\hline
\multirow{2}{*}{Method} & \multicolumn{3}{|c|}{Bedroom} & \multicolumn{3}{|c|}{Living room} & \multicolumn{3}{|c|}{Bathroom} & \multicolumn{3}{|c|}{Office}\\\cline{2-13}
  & Top1 & Top3 & Top5 & Top1 & Top3 & Top5 & Top1 & Top3 & Top5 & Top1 & Top3 & Top5 \\
\hline\hline
GRAINS [12] & 45.1 & 63.4 & 72.2 & 43.7 & 64.3 & 73.7 & 42.4 & 63.4 & 75.1 & 45.6 & 64.5 & 73.5 \\
Wang et al. [20] & 48.9 & 69.8 & 79.5 & 46.6 & 61.1 & 69.0 & 61.4 & 82.8 & 89.7 & 46.6 & 69.0 & 78.8 \\
\cline{1-13}
SGNet-tree & 60.0 & 78.1 & 85.4 & 63.7 & 80.7 & 87.2 & 61.3 & 83.4 & 90.4 & 59.6 & 76.5 & 84.9 \\
SGNet-sparse & 59.3 & 77.7 & 84.8 & 62.4 & 79.7 & 86.5 & 60.5 & 82.6 & 90.4 & 57.9 & 74.5 & 84.1 \\
SGNet-co-occur & 57.6 & 76.1 & 83.4 & 41.1 & 62.8 & 72.7 & 67.7 & 86.8 & 92.3 & 59.4 & 75.7 & 84.7 \\
SGNet-sum & 57.3 & 76.2 & 83.6 & 59.0 & 77.9 & 84.1 & 55.5 & 79.7 & 88.2 & 59.0 & 76.8 & 84.4 \\
SGNet-max & 63.1 & 79.6 & 86.1 & 61.1 & 79.3 & 85.3 & 67.7 & 88.2 & 92.9 & 60.3 & 78.9 & 85.0 \\
SGNet-vanilla-rnn & 64.3 & 81.3 & 87.8 & 65.0 & 81.1 & 87.3 & 67.6 & 86.5 & 92.0 & 62.3 & \textbf{80.4} & \textbf{86.8} \\
SGNet-no-attention & 60.2 & 77.6 & 83.8 & 62.2 & 79.8 & 86.4 & 61.6 & 82.1 & 89.7 & 57.1 & 76.8 & 83.6 \\
SGNet-dist-weights & 61.8 & 79.3 & 86.8 & 63.6 & 81.2 & 87.4 & 67.0 & 86.6 & 92.3 & 62.7 & 79.3 & 85.3 \\
\cline{1-13}
SceneGraphNet (full model) & \textbf{66.8} & \textbf{82.9} & \textbf{88.6} & \textbf{67.6} & \textbf{83.8} & \textbf{89.6} & \textbf{69.8} & \textbf{88.6} & \textbf{93.5} & \textbf{64.8} & 79.9 & 86.5 \\
\hline
\end{tabular}
\end{center}
\caption{Top-$K$ accuracy for category prediction for each room type.}
\label{table:scene_completion_with_others_for_each_room_type}
\end{table*}

\paragraph{Detailed results per room type.}

Table \ref{table:scene_completion_with_others_for_each_room_type} shows the  top-$K$ accuracy of category prediction for all different methods and our degraded variants per each room type ($K=1,3,5$) separately. Based on Figure \ref{fig:num_obj_distribution} and Table \ref{table:scene_completion_with_others_for_each_room_type}, we can see that for bathrooms that have low number of objects\ ($78\%$ of them has less than $20$ objects), our method has $8.4\%$ higher prediction accuracy than Wang et al. [20] in terms of top-$1$ accuracy, while for living rooms that contain more objects (only $15\%$ of them has less than $20$ objects), ours significantly outperforms Wang et al. [20] by a margin of $21.0\%$  in terms of top-$1$ accuracy.

\paragraph{Performance for different top-K accuracy.}
Figure \ref{fig:topk_performance} shows the top-$K$ accuracy of object category prediction over different $K$s ($K=1,2,...,10$) for competing methods.

\end{document}